\pdfoutput=1

\documentclass[11pt]{article}

\usepackage[]{acl}
\usepackage{hyperref}
\usepackage{times}
\usepackage{latexsym}
\usepackage{amsmath}
\usepackage{graphicx}
\usepackage{listings}
\usepackage[T1]{fontenc}

\usepackage[utf8]{inputenc}

\usepackage{microtype}

\title{Sentiment Analysis of Cyberbullying Data in Social Media }

\author{Pradeep Pujari$^{*}$ \\
  EleutherAI / USA \\
  \texttt{pkpujari@acm.org} \\\And
  Arvapalli Sai Susmitha$^{*}$ \\
  IIT Kanpur / India  \\  
  \texttt{arvapallisaisusmitha@gmail.com} \\
}

\begin{document}
\maketitle
\footnotetext{$^{*}$Equal contribution}
\begin{abstract}
\textbf{Social media has become an integral part of modern life, but it has also brought with it the pervasive issue of cyberbullying a serious menace in today's digital age. Cyberbullying, a form of harassment that occurs on social networks, has escalated alongside the growth of these platforms. Sentiment analysis holds significant potential not only for detecting bullying phrases but also for identifying victims who are at high risk of harm, whether to themselves or others. Our work focuses on leveraging deep learning and natural language understanding techniques to detect traces of bullying in social media posts. We developed a Recurrent Neural Network with Long Short-Term Memory (LSTM) cells, using different embeddings. One approach utilizes BERT embeddings, while the other replaces the embeddings layer with the recently released embeddings API from OpenAI. We conducted a performance comparison between these two approaches to evaluate their effectiveness in sentiment analysis of Formspring Cyberbullying data. Our Code is Available at : \href{https://github.com/ppujari/xcs224u}{https://github.com/ppujari/xcs224u}}

\end{abstract}
\section{Introduction}

In recent years, social media has become an integral part of daily life, facilitating communication and connection on an unprecedented scale. However, it has also contributed to a rise in cyberbullying harassment conducted online \citep{feinberg2009cyberbullying}. Unlike traditional bullying, cyberbullying can happen anytime, with online anonymity encouraging perpetrators to act without facing immediate consequences \citep{Hasan2023ARO}. This anonymity often leads to higher rates of cyberbullying compared to in-person bullying. Bullying is recognized as a major health issue by institutions like the American Academy of Pediatrics, has become especially concerning in educational settings \citep{xu2012learning}. The tragic case of Megan Meier, a young victim of harassment on MySpace, highlights the devastating impact of cyberbullying \cite{vines2015embedded}. Given the growing concern in society and workplaces, and the fact that studies show many students face cyberbullying \cite{li2006bullying, cross2008cyberbullying}, detecting and preventing cyberbullying promptly is essential.

In this work, we use Sentiment Analysis (SA) to classify texts as positive or negative with respect to cyberbullying. While traditional embeddings used in Recurrent Neural Networks (RNNs) are effective in capturing sequential patterns, they struggle with contextual dependencies in longer sentences and have limitations in understanding complex language structures \cite{elman1990finding,pascanu2013difficulty}.In contrast, we incorporate state-of-the-art Large Language Model (LLM) embeddings, specifically BERT embeddings \citep{kenton2019bert}, and use the OpenAI API to obtain embeddings that leverage self-attention mechanisms for sentiment analysis. These models excel in capturing intricate language features, making them highly effective for a range of downstream tasks \cite{radford2019language,raffel2020exploring}.
These embeddings offer a deeper contextual understanding compared to earlier models, allowing for a more nuanced detection of harmful content and a more precise evaluation of sentiment polarity. The task involves identifying sentences containing bullying tokens, assessing their polarity, and accurately classifying them into the relevant sentiment categories.


The development of such Natural Language Processing (NLP) algorithms, however, requires high-quality annotated data to measure performance accurately. Many popular Machine Learning (ML) techniques, especially Deep Neural Networks (DNNs), need large, annotated corpora to achieve high-quality classification results. Unfortunately, the availability of publicly accessible datasets for cyberbullying detection is limited. In this work, we use Kaggle’s Formspring data \cite{ppujari_2022_formspring,Reynolds2011UsingML} for Cyberbullying Detection to train and evaluate our models. We apply the advanced LLM-based techniques and assess the model’s performance in identifying cyberbullying.

In this work, we made the following contributions:
\begin{itemize}
    \item  Introduce two distinct hybrid methods for sentiment analysis in cyberbullying detection, leveraging advanced embeddings instead of traditional techniques. 
    \item Utilizes BERT embeddings with an RNN framework in one method and OpenAI embeddings with an RNN framework in another, specifically for sentiment analysis of cyberbullying data, aiding in cyberbullying detection. 
    \item Compare the performance of the proposed hybrid architectures using BERT and OpenAI embeddings for sentiment analysis on the Formspring dataset. 
\end{itemize}

\section{Literature review}
In the early stages of sentiment analysis, traditional methods such as rule-based and lexicon-based systems were widely used. Rule-based systems applied predefined linguistic rules to classify text, while lexicon-based methods relied on sentiment lexicons like AFINN and SentiWordNet\cite{baccianella2010sentiwordnet}, which assigned sentiment scores to individual words. These scores were then aggregated to determine the overall sentiment of a sentence or document. While these approaches were straightforward, they struggled to capture context-dependent meanings and often required extensive, manually curated lexicons. Similarly, pattern-based approaches like those described by \cite{Pang2008OpinionMA} utilized syntactic and semantic rules to detect sentiment-bearing structures within text, but they too required significant manual effort to maintain.

With the rise of machine learning, traditional feature engineering techniques became popular. Methods like TF-IDF (Term Frequency-Inverse Document Frequency) \cite{shi2011sentiment} transformed text data into meaningful features for sentiment classification, marking a shift from rule-based approaches. Classification algorithms such as Naive Bayes, Support Vector Machines (SVM), and Logistic Regression gained popularity in these early machine learning systems \cite{Joachims1999TextCW,McCallum1998ACO}. These methods improved flexibility but still faced challenges with complex sentiment detection and domain generalization. Feature selection and careful tuning were often necessary to optimize performance. 

Around the same time, distributed representation methods like Word2Vec\cite{Mikolov2013DistributedRO} and GloVe\cite{Pennington2014GloVeGV} were introduced, representing words in continuous vector spaces to capture their semantic relationships. Unlike TF-IDF's sparse features, word embeddings encode words based on context, enabling more effective sentiment classification. Word2Vec offered two model architectures, Continuous Bag-of-Words (CBOW) and Skip-Gram, both of which generated word embeddings by considering context words. GloVe took a different approach by using a global word co-occurrence matrix to create word embeddings, capable of capturing linear relationships within the vector space. \cite{Yin2009DetectionOH} utilized a traditional bag-of-words model with sentiment and contextual features to boost detection performance, while \cite{Nahar2014SemisupervisedLF} developed a semi-supervised fuzzy SVM approach incorporating sentiment and user metadata.

While these word embeddings revolutionized the way sentiment data was processed, they were still largely task-independent. In sentiment analysis, this posed challenges since word meanings could vary drastically depending on the context or domain. To address this, more advanced deep learning models, such as Recurrent Neural Networks (RNNs) and Long Short-Term Memory (LSTM) networks, were introduced to capture sequential dependencies in text \cite{Hochreiter1997LongSM}. This enabled better handling of long-range context, a critical factor in sentiment analysis.The paper \cite{Sahoo2023SentimentAU} discusses the use of RNNs, LSTMs, and transformer models in sentiment analysis, highlighting data preprocessing, feature extraction, and model architectures. Convolutional Neural Networks (CNNs) were also adapted to text processing, particularly excelling at capturing sentiment-bearing phrases and structures\cite{Kim2014ConvolutionalNN}. A study by \cite{Dang2020SentimentAB}, compares different deep learning models such as DNN, RNN, and CNN for sentiment analysis tasks, including sentiment polarity and aspect-based sentiment analysis. The paper \cite{Sharma2024ARO} explores the significance, challenges, and evolving methodologies in sentiment analysis, emphasizing the dynamic nature of the field. These studies collectively provide a robust understanding of how deep learning techniques can be leveraged for effective sentiment analysis.

Several studies have explored various approaches for cyberbullying detection, leveraging different models and techniques. \cite{Raj2021CyberbullyingDH} demonstrated the use of Bi-GRU and Bi-LSTM for high classification accuracy, while \cite{Dewani2021CyberbullyingDA} analyzed LSTM, BiLSTM, RNN, and GRU for detecting antisocial behavior. A hybrid neural network combining RNN for text and CNN for images was proposed by \cite{Agbaje2024NeuralNC} to detect cyberbullying from tweets. \cite{Iwendi2020CyberbullyingDS} applied four deep learning models Bidirectional LSTM (BLSTM), GRU, LSTM, and RNN for cyberbullying detection. The Hybrid Recurrent Residual Convolutional Neural Network (HRecRCNN) integrating sentiment analysis features was introduced by \cite{TT2021SentimentAA}. Sentiment analysis was also leveraged by \cite{Atoum2020CyberbullyingDT} using SVM and Naïve Bayes for harmful content classification, and later enhanced by \cite{Atoum2021CyberbullyingDN} with CNNs for improved feature extraction.

The introduction of transformer-based models like BERT and GPT revolutionized sentiment analysis and cyberbullying detection. BERT, introduced by \cite{Devlin2019BERTPO}, utilized bidirectional attention mechanisms to capture nuanced relationships between words in both forward and backward contexts, making it particularly effective for sentiment classification tasks. GPT, developed by \cite{Radford2018ImprovingLU}, employed a unidirectional architecture and excelled in tasks like text generation and understanding. These models significantly outperformed earlier techniques like Word2Vec and GloVe, thanks to their ability to capture deep contextual information. However, they also introduced challenges in terms of computational complexity and resource requirements \cite{Vaswani2017AttentionIA}.

The adoption of transformer-based models like BERT has not only advanced sentiment analysis but also cyberbullying detection, as these models provide superior text classification by understanding the subtle nuances of human language especially crucial for identifying abusive language on social media platforms. Our approach builds upon these advancements by leveraging two distinct hybrid architectures for sentiment analysis-based cyberbullying detection. One architecture integrates BERT embeddings(base model) with Recurrent Neural Networks (RNNs), while the other combines OpenAI API embeddings(text-embedding-3-small) with RNNs. Each hybrid model harnesses the rich contextual insights of its respective embedding method BERT or OpenAI while taking advantage of RNN’s ability to model the sequential nature of abusive language. This approach provides a comprehensive and robust framework for detecting cyberbullying across multiple posts.

\section{Formspring Dataset}


The \textbf{Formspring} dataset \cite{Reynolds2011UsingML}, curated by Kelly Reynolds and made available on \textit{Kaggle}\cite{ppujari_2022_formspring}, is one of the primary datasets used for sentiment analysis-based \textit{cyberbullying detection}. Formspring.me, a now-defunct question-and-answer-based social network, allowed users to post questions and comments anonymously, making it a significant source for studying online bullying behaviors. The dataset was collected through a crawl of the site in Summer 2010.

The dataset contains \textbf{12,772 samples} of user posts and answers, where each post was labeled as either containing cyberbullying or not. These labels were generated through \textit{Amazon Mechanical Turk}, with each post annotated by three workers. A post was classified as cyberbullying if at least two of the three annotators indicated the presence of bullying. Out of the 12,772 samples, \textbf{802 samples} (6.3\%) were labeled as cyberbullying, which mirrors the real-world imbalance between non-bullying and bullying content (with over \textbf{84\%} of the content labeled as non-bullying).\\ \textbf{Key fields} in the dataset include:
\begin{itemize}
    \item \textbf{Userid}: ID of the person responding to a post.
    \item \textbf{Asker}: ID of the person asking the question.
    \item \textbf{Post}: The complete question and answer string, separated by markers such as "Q:" and "A:". As part of preprocessing, these markers, along with any HTML encodings, were removed.
    \item \textbf{Ans\#, severity\#, bully\#}: The responses provided by Mechanical Turk workers. \textbf{Ans\#} refers to the answer given by the annotators, \textbf{severity\#} indicates a score from 0 to 10 based on the perceived intensity of the bullying, and \textbf{bully\#} identifies the specific words or phrases flagged as bullying.
\end{itemize}

The dataset contains approximately \textbf{300,000 tokens}. Harmful posts are slightly shorter on average, with \textbf{23 words per post}, compared to non-harmful posts, which have about \textbf{25 words}. Although the dataset is not particularly large by current machine learning standards, it remains a valuable source for studying cyberbullying due to its focus on anonymous interactions, which often encourage harmful behaviors.


The annotation process using \textit{Mechanical Turk} introduced some inconsistency in the labeling. Analysis suggests that \textbf{2.5\%} of posts classified as non-cyberbullying likely contain harmful content, and \textbf{15-20\%} of posts labeled as cyberbullying could be considered benign. Ideally, the dataset should be re-annotated for higher accuracy, but due to time constraints, the original labels were used in this work.\\

Preprocessing steps included:

\begin{itemize}
    \item \textbf{Handling Missing Values}: Replace 'None' values with \texttt{NaN} and map missing values in answer-related columns to 0 (indicating no bullying found).
    \item \textbf{Handling Corrupted User IDs}: Drop rows where \texttt{Userid} is corrupted and occurs fewer than 3 times.
    \item \textbf{Removal of HTML Tags}: Remove all HTML tags and encoded characters such as \texttt{<br>}.
    \item \textbf{Normalization of Repeated Characters}: Normalize words with exaggerated repetitions (e.g., \texttt{goooood} becomes \texttt{good}).
\end{itemize}

\noindent These are a few preprocessing steps that ensured cleaner data for model training. Furthermore, the dataset's target class was not annotated. \\

Logic to annotate the target as follows:


\begin{itemize}
    \item If two answers were identical and the severity score was greater than zero, we classified the data as 'Negative', indicating cyberbullying. If these conditions were not met, we classified the data as 'Positive', indicating non-bullying. 
    \item The dataset was then converted into numerical classes: No Bullying (Positive) = 1, Bullying (Negative) = -1.
    
\end{itemize}

\begin{figure}[ht]	
    \centering
    \includegraphics[width=0.5\textwidth]{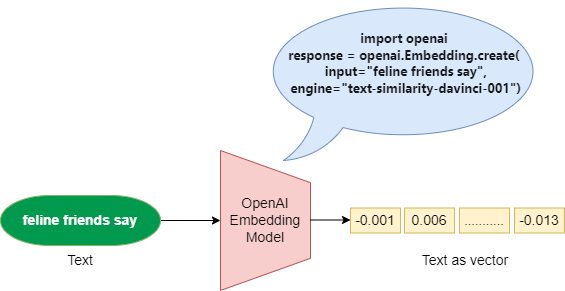}
    \caption{OpenAI API Embedding genration}
    \label{fig:Openai_embedding}
\end{figure}

\newpage
\section{Models}

\subsection{BERT Embeddings}
BERT (Bidirectional Encoder Representations from Transformers) embeddings are employed to extract contextual word representations. Unlike static word embeddings like \texttt{Word2Vec}, where each word has a fixed vector, BERT \cite{kenton2019bert} generates dynamic representations for each word based on its context. This allows for improved performance in NLP tasks, including cyberbullying detection. We used \textbf{BERT base} which has  12 encoder layers, 768 hidden units, 12 attention heads.\\
These transformer-based models outperform traditional methods by leveraging self-attention mechanisms and contextual embeddings.The attention mechanism is defined as $\text{Attention}(Q, K, V) = \text{softmax}\left(\frac{QK^T}{\sqrt{d_k}}\right) V$, where $Q$, $K$, and $V$ are query, key, and value matrices, and $d_k$ is the dimension of the key.





\subsection{OpenAI Embeddings}

The OpenAI embedding API introduces an endpoint that maps text and code to vector representations. We use text-embedding-3-small \cite{openai_embeddings_guide}, which are effective for tasks like clustering, topic modeling, and classification, outperforming many traditional models \cite{openai2023textcomparison}. These embeddings achieved a 20\% relative improvement in code search tasks \cite{neelakantan2022text}. The embeddings encode semantic relationships between words, phrases, or code snippets and can be obtained via a simple API request, as shown in Fig\ref{fig:Openai_embedding}. This encoding allows semantically similar inputs to cluster in vector space, making them useful for similarity search, grouping, or automated labeling.






\section{Experimentation}

\noindent\textbf{Algorithm: Sentiment Analysis using BERT or OpenAI Embeddings in RNN}
\begin{enumerate}
    \item \textbf{Input:} \\
    Input text \( T = \{t_1, t_2, \dots, t_n\} \), where \( t_i \) represents each token in the input sentence.

    \item \textbf{Step 1: Tokenization} \\
    Tokenize the input text \( T \) using:
    \begin{itemize}
        \item \textbf{BERT:} Use BERT’s tokenizer to obtain subword tokens \( T_{sub} = \{s_1, s_2, \dots, s_m\} \).
        \item \textbf{OpenAI:} Input the raw text \( T \) directly to OpenAI’s embedding API.
    \end{itemize}

    \item \textbf{Step 2: Embedding Extraction} \\
Extract contextual embeddings using:
    \begin{itemize}
        \item \textbf{BERT:} Pass tokenized text \( T_{sub} \) through a pre-trained BERT model to obtain embeddings \( E = \{e_1, e_2, \dots, e_m\} \).
        \item \textbf{OpenAI:} Use OpenAI's API to get high-dimensional vector embeddings \( E = \texttt{Embedding.create}(input=T) \).
    \end{itemize}

    \item \textbf{Step 3: Classification Model} \\
    Input the embeddings \( E \) into a RNN model to perform the classification task.

    \item \textbf{Step 4: Output} \\
    The final output is the predicted class \( y_{\text{pred}} \) with the highest probability.
\end{enumerate}
\hspace{1mm}

The model is trained using \textbf{Binary Cross Entropy Loss (BCELoss)}, which is suitable for binary classification tasks. \\

The BCELoss is defined as 
\[L =  \left( y \log(p) + (1 - y) \log(1 - p) \right)\]

 where $y$ is the true label and $p$ is the predicted probability for the positive class.\\

\begin{figure}[ht]	
    \centering
    \includegraphics[width=0.5\textwidth]{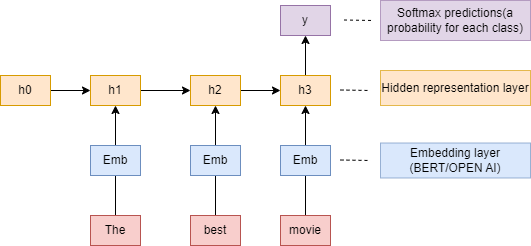}
    \caption{RNN network with BERT/OpenAI API embeddings}
    \label{fig:RNN_architecture}
\end{figure}

Incorporating BERT-base or OpenAI embeddings, i.e., our hybrid architectures, enhances the performance of the classifier compared to using only the RNN. This improvement is achieved by providing high-quality feature vectors extracted from the text data. Fig \ref{fig:RNN_architecture} shows the RNN network with BERT or OpenAI embeddings.\\
\newpage
\noindent\textbf{Implementation Details:}\\
The data was split into an 80:20 ratio for training and testing. We used a learning rate of 0.001 and applied gradient clipping to prevent exploding gradients, and dropout to mitigate overfitting. The Adam optimizer was employed for the training process, which was run for 10 epochs.
\section{Results}



We observed that the embedding representations, both BERT and OpenAI, were rich and dense with information, making them well-suited for the task of sentiment analysis in cyberbullying detection. However, OpenAI embeddings exhibited marginally better performance than BERT embeddings. 



The primary metrics used to evaluate model performance in this context were accuracy and macro F1 score. Accuracy is calculated as the ratio of correctly predicted instances to the total number of instances in the dataset, defined by the following formula:

\[
\text{Accuracy} = \frac{\text{Number of Correct Predictions}}{\text{Total Number of Predictions}} \times 100
\]

The macro F1 score provides an average F1 score across all classes, giving equal weight to each class. Since we have two classes, \( N = 2 \). The macro F1 score is calculated as follows:

\[
\text{Macro F1} = \frac{1}{N} \sum_{i=1}^{N} \frac{2 \times \text{Precision}_i \times \text{Recall}_i}{\text{Precision}_i + \text{Recall}_i}
\]
where \( \text{Precision}_i \) and \( \text{Recall}_i \) are the precision and recall for each class \( i \).
The results are summarized below:

\begin{table}[ht]
\centering
\begin{minipage}{0.47\textwidth}
\centering
\resizebox{\textwidth}{!}{%
\begin{tabular}{|c|c|c|}
\hline
\textbf{Model} & \textbf{Accuracy} & \textbf{Macro F1} \\
\hline
Basic RNN & 51\% & 38\% \\
BERT-base Hybrid (ours) & 56\% & 51\% \\
OpenAI Hybrid (ours) & 80\% & 79\% \\
\hline
\end{tabular}%
}
\caption{Results of different embeddings}
\end{minipage}
\end{table}

These results indicate that employing more advanced contextual embeddings, such as those from OpenAI, can lead to significant improvements in performance. The enhanced accuracy of OpenAI embeddings suggests that they are better at capturing the complexities and nuances of language in cyberbullying contexts, making them a valuable asset for future research and development in this area. Overall, this analysis underscores the importance of embedding choice in sentiment analysis tasks, especially in sensitive domains like cyberbullying detection.

\section{Conclusion}


An in-depth analysis of the Formspring dataset highlights significant opportunities for improving cyberbullying detection through advanced methodologies. Our study introduced a novel hybrid approach for sentiment analysis, integrating BERT and OpenAI embeddings within an RNN framework to address the complexities of cyberbullying data. 

Our findings revealed that both OpenAI and BERT embeddings significantly outperformed a basic RNN model in capturing the subtleties of cyberbullying language. Among them, OpenAI embeddings demonstrated the highest effectiveness, indicating their superior ability to understand context and sentiment in user-generated content. Looking ahead, the small size of the dataset poses challenges for model performance. To address this, we plan to explore zero-shot or few-shot learning techniques as future work, which could enhance the model's generalization and accuracy in predicting cyberbullying instances with limited labeled data. This direction promises to improve the robustness and applicability of detection systems in real-world scenarios.

\section{Acknowledgments}
We thank Prof. Christopher Potts, Petra Parikova, and the XCS224U course for inspiring this work, which started as part of the course.


\newpage
\newpage
\bibliography{custom}

\begin{thebibliography}{40}
\expandafter\ifx\csname natexlab\endcsname\relax\def\natexlab#1{#1}\fi

\bibitem[{Agbaje and Afolabi(2024)}]{Agbaje2024NeuralNC}
Michael Agbaje and Oreoluwa Afolabi. 2024.
\newblock \href {https://api.semanticscholar.org/CorpusID:270672750} {Neural network-based cyber-bullying and cyber-aggression detection using twitter(x) text}.
\newblock \emph{Revue d'Intelligence Artificielle}.

\bibitem[{Atoum(2020)}]{Atoum2020CyberbullyingDT}
Jalal~Omer Atoum. 2020.
\newblock \href {https://api.semanticscholar.org/CorpusID:235616725} {Cyberbullying detection through sentiment analysis}.
\newblock \emph{2020 International Conference on Computational Science and Computational Intelligence (CSCI)}, pages 292--297.

\bibitem[{Atoum(2021)}]{Atoum2021CyberbullyingDN}
Jalal~Omer Atoum. 2021.
\newblock \href {https://api.semanticscholar.org/CorpusID:249928806} {Cyberbullying detection neural networks using sentiment analysis}.
\newblock \emph{2021 International Conference on Computational Science and Computational Intelligence (CSCI)}, pages 158--164.

\bibitem[{Baccianella et~al.(2010)Baccianella, Esuli, Sebastiani et~al.}]{baccianella2010sentiwordnet}
Stefano Baccianella, Andrea Esuli, Fabrizio Sebastiani, et~al. 2010.
\newblock Sentiwordnet 3.0: an enhanced lexical resource for sentiment analysis and opinion mining.
\newblock In \emph{Lrec}, volume~10, pages 2200--2204. Valletta.

\bibitem[{Cross(2008)}]{cross2008cyberbullying}
D~Cross. 2008.
\newblock Cyberbullying: International comparisons, implications, and recommendations.
\newblock In \emph{20th biennial meeting of the International Society for the Study of Behavioural Development, Wurzburg, Germany}.

\bibitem[{Dang et~al.(2020)Dang, Garc{\'i}a, and de~la Prieta}]{Dang2020SentimentAB}
Nhan~Cach Dang, Mar{\'i}a N.~Moreno Garc{\'i}a, and Fernando de~la Prieta. 2020.
\newblock \href {https://api.semanticscholar.org/CorpusID:216209052} {Sentiment analysis based on deep learning: A comparative study}.
\newblock \emph{ArXiv}, abs/2006.03541.

\bibitem[{Devlin et~al.(2019)Devlin, Chang, Lee, and Toutanova}]{Devlin2019BERTPO}
Jacob Devlin, Ming-Wei Chang, Kenton Lee, and Kristina Toutanova. 2019.
\newblock \href {https://api.semanticscholar.org/CorpusID:52967399} {Bert: Pre-training of deep bidirectional transformers for language understanding}.
\newblock In \emph{North American Chapter of the Association for Computational Linguistics}.

\bibitem[{Dewani et~al.(2021)Dewani, Memon, and Bhatti}]{Dewani2021CyberbullyingDA}
Amirita Dewani, Mohsin~Ali Memon, and Sania Bhatti. 2021.
\newblock \href {https://api.semanticscholar.org/CorpusID:245425394} {Cyberbullying detection: advanced preprocessing techniques \& deep learning architecture for roman urdu data}.
\newblock \emph{Journal of Big Data}, 8.

\bibitem[{Elman(1990)}]{elman1990finding}
Jeffrey~L Elman. 1990.
\newblock Finding structure in time.
\newblock \emph{Cognitive science}, 14(2):179--211.

\bibitem[{Feinberg and Robey(2009)}]{feinberg2009cyberbullying}
Ted Feinberg and Nicole Robey. 2009.
\newblock Cyberbullying.
\newblock \emph{The education digest}, 74(7):26.

\bibitem[{Hasan et~al.(2023)Hasan, Hossain, Mukta, Akter, Ahmed, and Islam}]{Hasan2023ARO}
Md.~Tarek Hasan, Md~Al~Emran Hossain, Md. Saddam~Hossain Mukta, Arifa Akter, Mohiuddin Ahmed, and Salekul Islam. 2023.
\newblock \href {https://api.semanticscholar.org/CorpusID:258668089} {A review on deep-learning-based cyberbullying detection}.
\newblock \emph{Future Internet}, 15:179.

\bibitem[{Hochreiter and Schmidhuber(1997)}]{Hochreiter1997LongSM}
Sepp Hochreiter and J{\"u}rgen Schmidhuber. 1997.
\newblock \href {https://api.semanticscholar.org/CorpusID:1915014} {Long short-term memory}.
\newblock \emph{Neural Computation}, 9:1735--1780.

\bibitem[{Iwendi et~al.(2020)Iwendi, Srivastava, Khan, and Maddikunta}]{Iwendi2020CyberbullyingDS}
Celestine Iwendi, Gautam Srivastava, Suleman Khan, and Praveen Kumar~Reddy Maddikunta. 2020.
\newblock \href {https://api.semanticscholar.org/CorpusID:225127474} {Cyberbullying detection solutions based on deep learning architectures}.
\newblock \emph{Multimedia Systems}, 29:1839--1852.

\bibitem[{Joachims(1999)}]{Joachims1999TextCW}
Thorsten Joachims. 1999.
\newblock \href {https://api.semanticscholar.org/CorpusID:2427083} {Text categorization with support vector machines: Learning with many relevant features}.
\newblock In \emph{European Conference on Machine Learning}.

\bibitem[{Kenton and Toutanova(2019)}]{kenton2019bert}
Jacob Devlin Ming-Wei~Chang Kenton and Lee~Kristina Toutanova. 2019.
\newblock Bert: Pre-training of deep bidirectional transformers for language understanding.
\newblock In \emph{Proceedings of naacL-HLT}, volume~1, page~2.

\bibitem[{Kim(2014)}]{Kim2014ConvolutionalNN}
Yoon Kim. 2014.
\newblock \href {https://api.semanticscholar.org/CorpusID:9672033} {Convolutional neural networks for sentence classification}.
\newblock In \emph{Conference on Empirical Methods in Natural Language Processing}.

\bibitem[{Li(2006)}]{li2006bullying}
Q~Li. 2006.
\newblock Bullying, cyberbullying, and victimization in canada.
\newblock In \emph{annual conference of the Canadian Society for the Study of Education, Toronto, Ontario, Canada}.

\bibitem[{McCallum and Nigam(1998)}]{McCallum1998ACO}
Andrew McCallum and Kamal Nigam. 1998.
\newblock \href {https://api.semanticscholar.org/CorpusID:7311285} {A comparison of event models for naive bayes text classification}.
\newblock In \emph{AAAI Conference on Artificial Intelligence}.

\bibitem[{Mikolov et~al.(2013)Mikolov, Sutskever, Chen, Corrado, and Dean}]{Mikolov2013DistributedRO}
Tomas Mikolov, Ilya Sutskever, Kai Chen, Gregory~S. Corrado, and Jeffrey Dean. 2013.
\newblock \href {https://api.semanticscholar.org/CorpusID:16447573} {Distributed representations of words and phrases and their compositionality}.
\newblock In \emph{Neural Information Processing Systems}.

\bibitem[{Nahar et~al.(2014)Nahar, Al-Maskari, Li, and Pang}]{Nahar2014SemisupervisedLF}
Vinita Nahar, Sanad Al-Maskari, Xue Li, and Chaoyi Pang. 2014.
\newblock \href {https://api.semanticscholar.org/CorpusID:46718864} {Semi-supervised learning for cyberbullying detection in social networks}.
\newblock In \emph{Australasian Database Conference}.

\bibitem[{Neelakantan et~al.(2022)Neelakantan, Xu, Puri, Radford, Han, Tworek, Yuan, Tezak, Kim, Hallacy et~al.}]{neelakantan2022text}
Arvind Neelakantan, Tao Xu, Raul Puri, Alec Radford, Jesse~Michael Han, Jerry Tworek, Qiming Yuan, Nikolas Tezak, Jong~Wook Kim, Chris Hallacy, et~al. 2022.
\newblock Text and code embeddings by contrastive pre-training.
\newblock \emph{arXiv preprint arXiv:2201.10005}.

\bibitem[{OpenAI(2023)}]{openai2023textcomparison}
OpenAI. 2023.
\newblock \href {https://cookbook.openai.com/articles/text_comparison_examples} {Text comparison examples}.
\newblock Accessed: 2024-11-07.

\bibitem[{OpenAI(2024)}]{openai_embeddings_guide}
OpenAI. 2024.
\newblock Openai embeddings guide.
\newblock \url{https://platform.openai.com/docs/guides/embeddings}.
\newblock Accessed: 2024-03-07.

\bibitem[{Pang and Lee(2008)}]{Pang2008OpinionMA}
Bo~Pang and Lillian Lee. 2008.
\newblock \href {https://api.semanticscholar.org/CorpusID:207178694} {Opinion mining and sentiment analysis}.
\newblock \emph{Found. Trends Inf. Retr.}, 2:1--135.

\bibitem[{Pascanu(2013)}]{pascanu2013difficulty}
R~Pascanu. 2013.
\newblock On the difficulty of training recurrent neural networks.
\newblock \emph{arXiv preprint arXiv:1211.5063}.

\bibitem[{Pennington et~al.(2014)Pennington, Socher, and Manning}]{Pennington2014GloVeGV}
Jeffrey Pennington, Richard Socher, and Christopher~D. Manning. 2014.
\newblock \href {https://api.semanticscholar.org/CorpusID:1957433} {Glove: Global vectors for word representation}.
\newblock In \emph{Conference on Empirical Methods in Natural Language Processing}.

\bibitem[{Pujari(2022)}]{ppujari_2022_formspring}
Pratik Pujari. 2022.
\newblock \href {https://www.kaggle.com/datasets/ppujari/formspring-csv/data} {Formspring dataset for cyberbullying detection}.
\newblock Accessed: 2022-11-06.

\bibitem[{Radford et~al.(2018)Radford, Narasimhan, Salimans, and Sutskever}]{Radford2018ImprovingLU}
Alec Radford, Karthik Narasimhan, Tim Salimans, and Ilya Sutskever. 2018.
\newblock Improving language understanding by generative pre-training.

\bibitem[{Radford et~al.(2019)Radford, Wu, Child, Luan, Amodei, Sutskever et~al.}]{radford2019language}
Alec Radford, Jeffrey Wu, Rewon Child, David Luan, Dario Amodei, Ilya Sutskever, et~al. 2019.
\newblock Language models are unsupervised multitask learners.
\newblock \emph{OpenAI blog}, 1(8):9.

\bibitem[{Raffel et~al.(2020)Raffel, Shazeer, Roberts, Lee, Narang, Matena, Zhou, Li, and Liu}]{raffel2020exploring}
Colin Raffel, Noam Shazeer, Adam Roberts, Katherine Lee, Sharan Narang, Michael Matena, Yanqi Zhou, Wei Li, and Peter~J Liu. 2020.
\newblock Exploring the limits of transfer learning with a unified text-to-text transformer.
\newblock \emph{Journal of machine learning research}, 21(140):1--67.

\bibitem[{Raj et~al.(2021)Raj, Agarwal, Bharathy, Narayan, and Prasad}]{Raj2021CyberbullyingDH}
Chahat Raj, Ayush Agarwal, Gnana Bharathy, Bhuva Narayan, and Mukesh Prasad. 2021.
\newblock \href {https://api.semanticscholar.org/CorpusID:244299913} {Cyberbullying detection: Hybrid models based on machine learning and natural language processing techniques}.
\newblock \emph{Electronics}.

\bibitem[{Reynolds et~al.(2011)Reynolds, Kontostathis, and Edwards}]{Reynolds2011UsingML}
Kelly Reynolds, April Kontostathis, and Lynne Edwards. 2011.
\newblock \href {https://api.semanticscholar.org/CorpusID:9745068} {Using machine learning to detect cyberbullying}.
\newblock \emph{2011 10th International Conference on Machine Learning and Applications and Workshops}, 2:241--244.

\bibitem[{Sahoo et~al.(2023)Sahoo, Wankhade, and Singh}]{Sahoo2023SentimentAU}
Chinmayee Sahoo, Mayuri Wankhade, and Binod~Kumar Singh. 2023.
\newblock \href {https://api.semanticscholar.org/CorpusID:265353039} {Sentiment analysis using deep learning techniques: a comprehensive review}.
\newblock \emph{International Journal of Multimedia Information Retrieval}, 12:1--23.

\bibitem[{Sharma et~al.(2024)Sharma, Ali, and Kabir}]{Sharma2024ARO}
Neeraj~Anand Sharma, A.~B. M.~Shawkat Ali, and Muhammad~Ashad Kabir. 2024.
\newblock \href {https://api.semanticscholar.org/CorpusID:270907664} {A review of sentiment analysis: tasks, applications, and deep learning techniques}.
\newblock \emph{International Journal of Data Science and Analytics}.

\bibitem[{Shi and Li(2011)}]{shi2011sentiment}
Han-Xiao Shi and Xiao-Jun Li. 2011.
\newblock A sentiment analysis model for hotel reviews based on supervised learning.
\newblock In \emph{2011 International Conference on Machine Learning and Cybernetics}, volume~3, pages 950--954. IEEE.

\bibitem[{T.T and Jeetha(2021)}]{TT2021SentimentAA}
Sherly T.T and B.~Rosiline Jeetha. 2021.
\newblock \href {https://api.semanticscholar.org/CorpusID:242191735} {Sentiment analysis and deep learning based cyber bullying detection in twitter dataset}.
\newblock \emph{International Journal of Recent Technology and Engineering (IJRTE)}.

\bibitem[{Vaswani et~al.(2017)Vaswani, Shazeer, Parmar, Uszkoreit, Jones, Gomez, Kaiser, and Polosukhin}]{Vaswani2017AttentionIA}
Ashish Vaswani, Noam~M. Shazeer, Niki Parmar, Jakob Uszkoreit, Llion Jones, Aidan~N. Gomez, Lukasz Kaiser, and Illia Polosukhin. 2017.
\newblock \href {https://api.semanticscholar.org/CorpusID:13756489} {Attention is all you need}.
\newblock In \emph{Neural Information Processing Systems}.

\bibitem[{Vines(2015)}]{vines2015embedded}
James~E Vines. 2015.
\newblock \emph{An embedded case study of the proposed Megan Meier Cyberbullying Prevention statute \& the proposed Tyler Clementi Higher Education Anti-Harassment statute}.
\newblock Ph.D. thesis, Clemson University.

\bibitem[{Xu et~al.(2012)Xu, Jun, Zhu, and Bellmore}]{xu2012learning}
Jun-Ming Xu, Kwang-Sung Jun, Xiaojin Zhu, and Amy Bellmore. 2012.
\newblock Learning from bullying traces in social media.
\newblock In \emph{Proceedings of the 2012 conference of the North American chapter of the association for computational linguistics: Human language technologies}, pages 656--666.

\bibitem[{Yin et~al.(2009)Yin, Xue, Hong, Davison, Kontostathis, Edwards et~al.}]{Yin2009DetectionOH}
Dawei Yin, Zhenzhen Xue, Liangjie Hong, Brian~D Davison, April Kontostathis, Lynne Edwards, et~al. 2009.
\newblock Detection of harassment on web 2.0.
\newblock \emph{Proceedings of the Content Analysis in the WEB}, 2(0):1--7.

\end{thebibliography}




\end{document}